%% file: neurips_2026.tex
\documentclass{article}
\PassOptionsToPackage{numbers, compress}{natbib}

\usepackage[preprint]{neurips_2026}

\usepackage[utf8]{inputenc} 
\usepackage[T1]{fontenc}    
\usepackage{hyperref}       
\usepackage{url}            
\usepackage{booktabs}       
\usepackage{amsfonts}       
\usepackage{nicefrac}       
\usepackage{microtype}      
\usepackage{xcolor}         
\usepackage{subcaption}
\usepackage{graphicx}
\usepackage{amsmath}
\usepackage{amssymb}
\usepackage[ruled,vlined]{algorithm2e}

\definecolor{coral}{RGB}{255,127,80}

\title{Adaptive Greedy Frame Selection for \\ Long-Video Understanding}

\author{%
  Yuning Huang \\
  Purdue University\\
  West Lafayette, IN 47906 \\
  \And
  Xiaoyu Ji \\
  Purdue University\\
  West Lafayette, IN 47906 \\
  \And
  Joseph Huang \\
  Purdue University\\
  West Lafayette, IN 47906 \\
  \And
  Yichi Zhang \\
  Purdue University\\
  West Lafayette, IN 47906 \\
  \And
  Fengqing Zhu \\
  Purdue University\\
  West Lafayette, IN 47906 \\
}

\begin{document}

\maketitle

\begin{abstract}
Large vision--language models (VLMs) are increasingly used for long-video question answering, but inference remains constrained by the number of input frames and visual tokens. Most frame-selection pipelines, however, apply a single sampling rule to all questions. We challenge this assumption and show that different question types benefit from different relevance--coverage trade-offs. We formulate frame selection as query-aware subset selection over a bounded 1~FPS candidate pool capped at 1,000 frames. Candidate frames are embedded in two complementary spaces: SigLIP for question relevance and DINOv2 for semantic similarity. We greedily optimize a weighted sum of a modular relevance term and a facility-location coverage term; the resulting surrogate is normalized, monotone, and submodular, yielding the standard $(1-1/e)$ approximation guarantee under a fixed frame budget. This formulation yields four interpretable presets, ranging from relevance-only to coverage-only selection. Experiments on MLVU and LongVideoBench with Qwen2-VL and Qwen3-VL show that relevance--coverage strategies improve over uniform sampling and, in most settings, over Adaptive Keyframe Sampling (AKS), especially at small frame budgets. A post-hoc category oracle quantifies the headroom from perfect category-aware preset selection, while a deployable MLVU router based on a lightweight text-only question-type classifier improves over both AKS and the best fixed preset on a held-out test split. These results support question-adaptive frame selection as a practical way to use limited visual-token budgets more effectively.
\end{abstract}

\input{section/1_intro}
\input{section/2_related}
\input{section/3_method}
\input{section/4_exp}
\input{section/5_conclusion}

\bibliographystyle{splncs04}
\bibliography{main}


\appendix

\newpage
\begin{algorithm}[t]
\caption{Greedy Relevance + Facility-Location Coverage Selection}
\label{alg:greedy}
\KwIn{SigLIP frame embeddings $\{\mathbf{v}_i\}_{i=1}^N$ (normalized), SigLIP text embedding $\mathbf{t}$ (normalized); DINOv2 embeddings $\{\mathbf{d}_i\}_{i=1}^N$ (normalized); subset size $K$; weights $\alpha,\beta$}
\KwOut{selected positions $S$ (sorted)}
Compute relevance $r_i \leftarrow \max(\mathbf{v}_i^\top \mathbf{t}, 0)$ for all $i$\;
Compute similarity matrix $s_{j,i} \leftarrow \mathbf{d}_j^\top \mathbf{d}_i$\;
Initialize $S\leftarrow \emptyset$, and $c_j \leftarrow b~(=-1)$ for all $j$\;
\For{$\ell=1$ \KwTo $\min(K,N)$}{
    \ForEach{candidate $i\notin S$}{
        $\Delta_C(i\mid S) \leftarrow \frac{1}{N}\sum_{j=1}^N \left[\max(c_j, s_{j,i}) - c_j\right]$\;
        $\Delta(i\mid S)\leftarrow \alpha r_i + \beta \Delta_C(i\mid S)$\;
    }
    $i^\star \leftarrow \arg\max_{i\notin S} \Delta(i\mid S)$\;
    $S\leftarrow S\cup\{i^\star\}$\;
    $c_j \leftarrow \max(c_j, s_{j,i^\star})$ for all $j$\;
}
\Return{$\text{sort}(S)$}\;
\end{algorithm}

\section{Additional Analysis of Question-Type Routing}
\label{app:qtype-routing}

This appendix provides additional evidence for the question-type adaptive selection design in
Sec.~\ref{sec:qtype-method}. We first evaluate whether MLVU question types can be reliably
predicted from question text alone, and then examine how different relevance--coverage presets
behave across question categories.

\subsection{Question-type classifier}
\label{app:qtype-classifier}

\begin{figure}[t]
    \centering
    \includegraphics[width=0.6\linewidth]{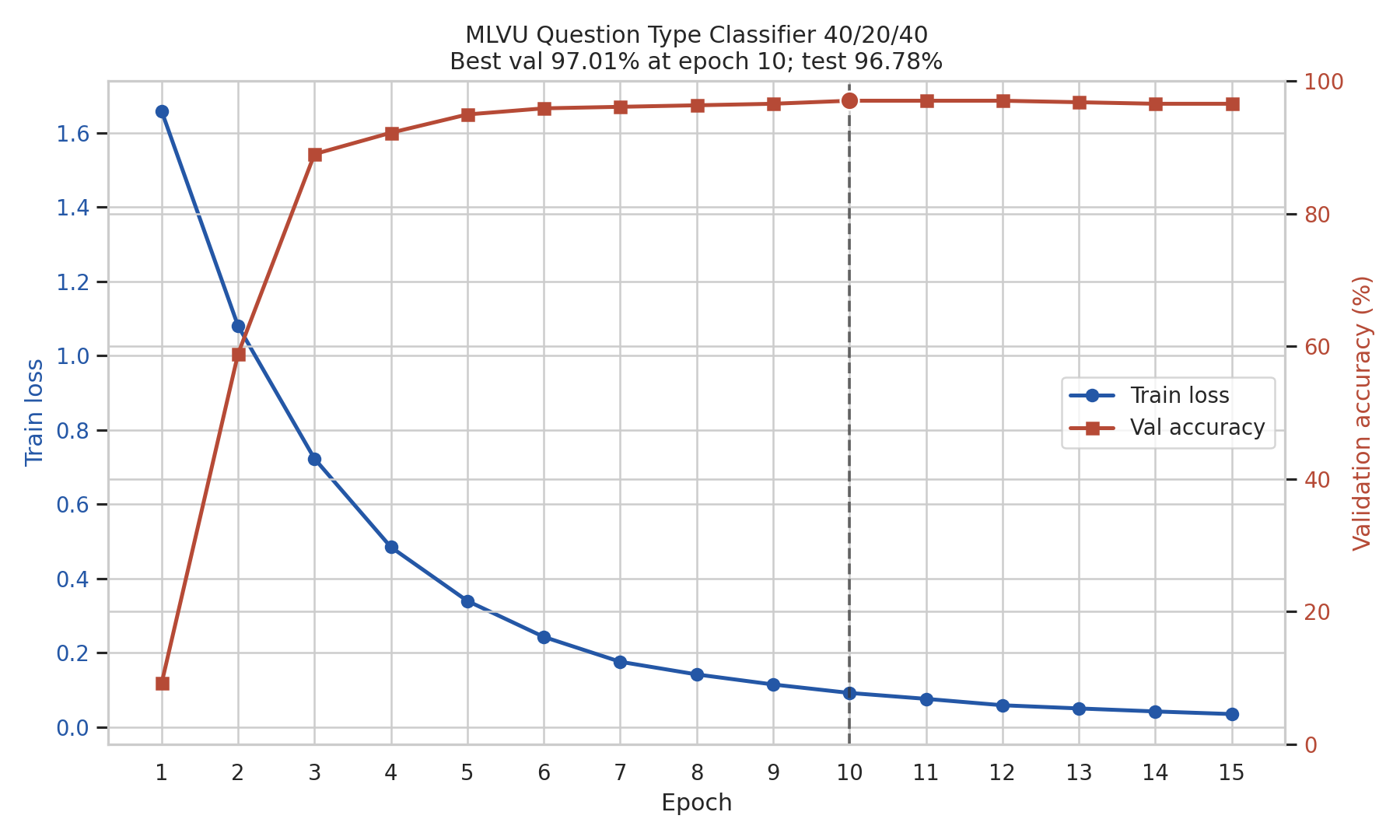}
    \caption{\textbf{Training dynamics of the MLVU question-type classifier.}
    The text-only classifier is trained on the 40/20/40 MLVU split used for the deployable
    adaptive routing experiment. The best validation accuracy is $97.01\%$ at epoch 10,
    with held-out test accuracy $96.78\%$.}
    \label{fig:qtype_classifier_training}
\end{figure}

The deployable adaptive strategy uses a lightweight text-only classifier to predict the MLVU
question type before frame selection. Fig.~\ref{fig:qtype_classifier_training} shows that the
classifier converges rapidly under the 40/20/40 train/validation/test split. Validation accuracy
reaches $97.01\%$ at epoch 10, and the corresponding held-out test accuracy is $96.78\%$.
This high test accuracy indicates that the routing signal used by the deployable strategy can be
obtained from the question text alone, without using video frames, ground-truth answers, or
test-set performance.

\begin{figure}[t]
    \centering
    \includegraphics[width=0.6\linewidth]{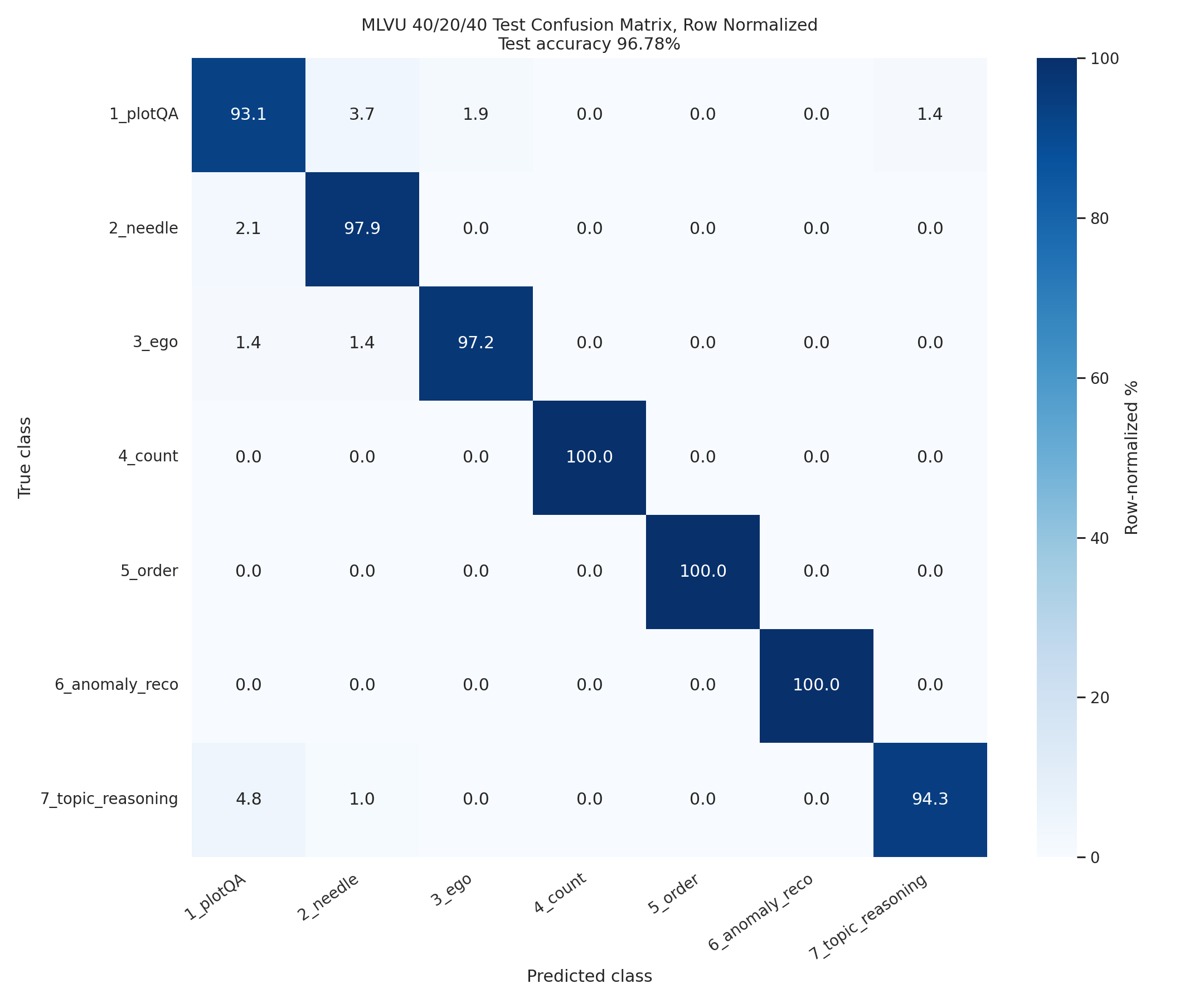}
    \caption{\textbf{Held-out MLVU question-type confusion matrix.}
    Rows are ground-truth MLVU question types and columns are predicted types. Values are
    row-normalized percentages. The classifier achieves $96.78\%$ test accuracy, with most
    mistakes occurring between semantically related question types such as \texttt{plotQA}
    and \texttt{topic\_reasoning}.}
    \label{fig:qtype_classifier_confusion}
\end{figure}

Fig.~\ref{fig:qtype_classifier_confusion} reports the row-normalized confusion matrix on the
held-out test split. Most categories are classified with high accuracy: \texttt{count},
\texttt{order}, and \texttt{anomaly\_reco} reach $100.0\%$ accuracy, while \texttt{needle}
and \texttt{ego} reach $97.9\%$ and $97.2\%$, respectively. The remaining errors are
concentrated in semantically adjacent categories. For example, a small fraction of
\texttt{plotQA} questions are predicted as \texttt{needle} or \texttt{ego}, and a small
fraction of \texttt{topic\_reasoning} questions are predicted as \texttt{plotQA}. These
confusions are expected because some plot-level and topic-level questions share similar
surface forms. Overall, the confusion matrix supports the feasibility of using predicted
question type as a practical routing variable.

\subsection{Per-category behavior of relevance--coverage presets}
\label{app:per-category-presets}

\begin{figure}[t]
    \centering
    \includegraphics[width=0.9\linewidth]{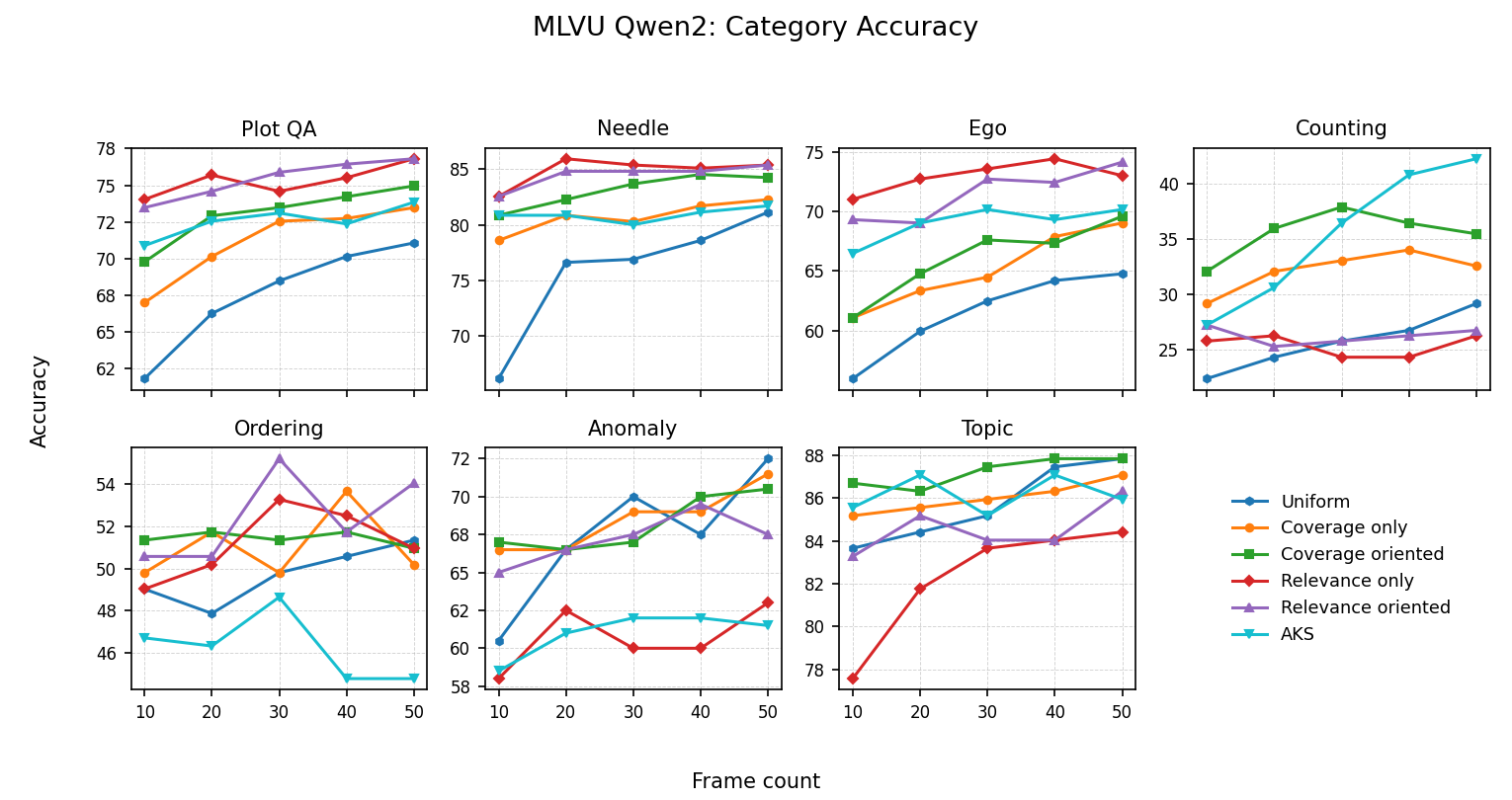}
    \caption{\textbf{Per-category MLVU accuracy for Qwen2-VL.}
    Each panel shows one MLVU question type across frame budgets. The results reveal
    heterogeneous strategy preferences: relevance-oriented selection is strong for several
    localized-evidence categories, while coverage-oriented selection or AKS is more effective
    for categories requiring broader aggregation, such as counting and topic reasoning.}
    \label{fig:mlvu_category_qwen2}
\end{figure}

\begin{figure}[t]
    \centering
    \includegraphics[width=0.9\linewidth]{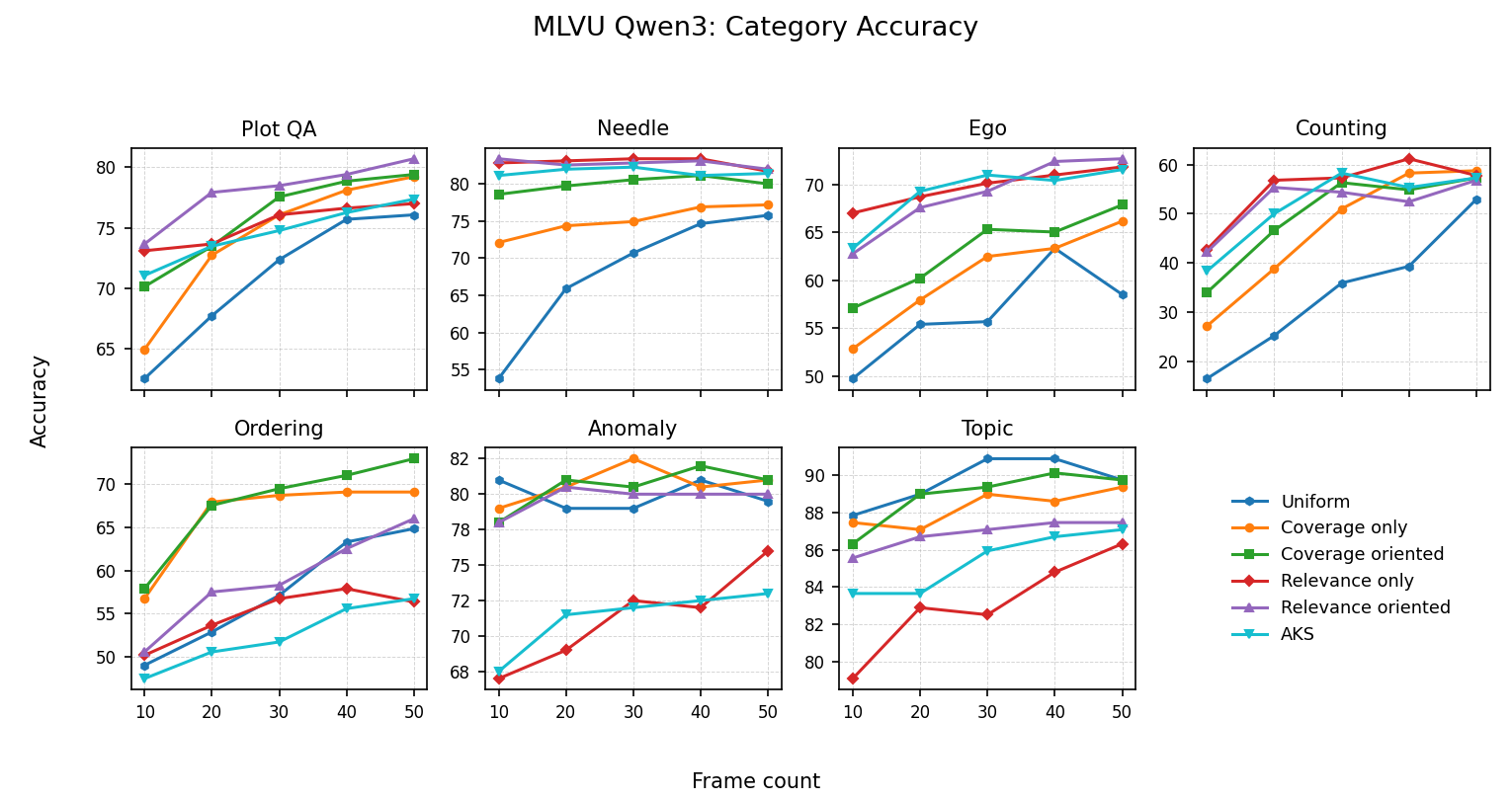}
    \caption{\textbf{Per-category MLVU accuracy for Qwen3-VL.}
    The stronger Qwen3-VL backbone changes some category-level preferences, but the overall
    pattern remains: different question types favor different relevance--coverage trade-offs.
    Coverage-oriented selection is particularly effective for categories requiring broad temporal
    or semantic context, while relevance-heavy strategies are strong for several localized-evidence
    categories.}
    \label{fig:mlvu_category_qwen3}
\end{figure}

Fig.~\ref{fig:mlvu_category_qwen2} and Fig.~\ref{fig:mlvu_category_qwen3} show per-category
MLVU accuracy as a function of the frame budget for Qwen2-VL and Qwen3-VL\@. These plots
provide a finer-grained view of the aggregate results in Table~\ref{tab:main_results}. The key
observation is that no single fixed relevance--coverage preset dominates across all categories.
Instead, the preferred trade-off depends on both the question type and the VLM backbone.

For Qwen2-VL, relevance-heavy strategies are strong for categories that often require localized
evidence, such as \texttt{plotQA}, \texttt{needle}, and \texttt{ego}. In contrast, coverage-heavy
or broader-selection strategies are more useful for categories such as \texttt{count},
\texttt{anomaly\_reco}, and \texttt{topic\_reasoning}, where the answer may depend on
aggregating evidence across multiple moments or maintaining broad video context. The
\texttt{count} category is a particularly clear example: pure relevance strategies perform poorly,
whereas AKS and coverage-oriented selection improve substantially as the frame budget grows.

For Qwen3-VL, the same qualitative heterogeneity remains, although the stronger backbone shifts
some category preferences. Relevance-oriented or relevance-only selection performs well for
\texttt{plotQA}, \texttt{needle}, \texttt{ego}, and \texttt{count}, while coverage-oriented
selection is especially strong for \texttt{order}, \texttt{anomaly\_reco}, and
\texttt{topic\_reasoning}. This indicates that the relevance--coverage trade-off is not only a
property of the dataset category, but can also depend on the downstream VLM\@. Nevertheless, the
main conclusion is consistent across both backbones: mixed relevance--coverage strategies are
often more robust than pure relevance or pure coverage, and category-aware routing can exploit
the fact that different question types prefer different selectors.

Together, these results explain why the deployable adaptive strategy improves over the strongest
fixed strategy in Table~\ref{tab:deployable_results}. The classifier results show that question
types can be predicted accurately before inference, and the per-category curves show that those
types have meaningfully different selector preferences. Thus, the adaptive policy does not rely on
post-hoc test information: it uses a predictable question attribute to choose among fixed
relevance--coverage presets selected on validation data.


\newpage
\newpage
\input{checklist.tex}

\end{document}

%% file: section/1_intro.tex
\section{Introduction}
\label{sec:intro}

Large vision--language models (VLMs) are increasingly applied to video question answering and multimodal reasoning, but long videos remain difficult to process under fixed context and compute budgets. Feeding more frames quickly increases visual-token cost, whereas sparse sampling can miss rare but decisive moments. Recent work addresses this bottleneck with memory mechanisms and sparse representations~\cite{Song_2024_CVPR}, hierarchical or adaptive video representations~\cite{Wang_2025_CVPR}, retrieval-augmented pipelines~\cite{Ma_2025_CVPR}, and training-free or lightweight frame-selection heuristics~\cite{Liu_2025_CVPR}. In parallel, learnable selection policies and optimization-based samplers have been proposed to choose informative frames under strict budgets~\cite{Buch_2025_CVPR,Tang_2025_CVPR,yao-etal-2025-generative}. Benchmarks such as Video-MME and LongVideoBench further show that long-video performance depends strongly on how well a model allocates limited visual tokens to relevant temporal evidence~\cite{Fu_2025_CVPR,wu2024longvideobench}.

Frame selection, however, is not merely a relevance-ranking problem. In practice, two failure modes often appear. A selector may suffer from \emph{redundancy collapse}, concentrating many frames around the same salient segment and wasting budget on near-duplicates. Conversely, it may suffer from \emph{coverage collapse}, spreading too uniformly and missing localized evidence needed to answer a specific question. These failures motivate a relevance--coverage view of long-video frame selection: selected frames should align with the question while also representing the video's broader semantic structure. This perspective connects to classic relevance--diversity retrieval, such as Maximal Marginal Relevance~\cite{Carbonell_1998_MMR}, and to submodular coverage objectives that admit efficient greedy maximization with approximation guarantees~\cite{nemhauser1978submodular,Krause_2014_Submodular,lin-bilmes-2011-class}. Diversity can also be promoted by DPP/log-determinant objectives~\cite{kulesza2012dpp,Zhang_2025_AdaRDKey}.

Our key observation is that the best relevance--coverage trade-off is \emph{question dependent}. Summary-style questions often benefit from broad temporal and semantic coverage, whereas fine-grained event questions may require stronger query relevance with only mild redundancy control. A single fixed selection rule is therefore unlikely to be optimal across heterogeneous question types. We address this with a query-adaptive greedy frame selector that (i) optimizes a parameterized relevance--coverage objective and (ii) routes questions to one of a small number of interpretable trade-off presets using a lightweight question-type classifier. This design is complementary to recent adaptive acquisition frameworks that decide when and how to gather additional video evidence~\cite{Zou_2026_VideoBrain}.

We evaluate on MLVU~\cite{Zhou_2025_CVPR} and LongVideoBench~\cite{wu2024longvideobench}. MLVU is central to our study because its question taxonomy enables controlled analysis of type-conditioned frame selection and a held-out deployable routing experiment. LongVideoBench provides a complementary benchmark-level check that the same relevance--coverage trends persist beyond MLVU\@. Because deployable routing requires category supervision for training and validation, the end-to-end adaptive-routing result is reported on MLVU, while LongVideoBench is used for fixed-strategy and oracle-headroom analysis.

\paragraph{Contributions.}
Our main contributions are:
(1) a query-aware greedy frame-selection objective that combines SigLIP-based modular relevance with DINOv2-based facility-location coverage under a fixed frame budget, yielding a normalized monotone submodular surrogate with the standard $(1-1/e)$ greedy approximation guarantee;
(2) an interpretable question-adaptive selection framework that separates a non-deployable category oracle from a deployable validation-selected router over four relevance--coverage presets; and
(3) experiments on MLVU and LongVideoBench with Qwen2-VL and Qwen3-VL showing that relevance--coverage selection improves low-budget frame selection, that no single preset dominates across categories and budgets, and that a held-out MLVU deployable router improves over both AKS and the best fixed strategy.

%% file: section/2_related.tex
\section{Related Work}
\label{sec:related}

\subsection{Long-video understanding benchmarks and question taxonomies}
Long-video understanding benchmarks stress-test VLMs under long temporal horizons and tight context budgets. MLVU~\cite{Zhou_2025_CVPR} is particularly suitable for studying question-adaptive frame selection because it provides structured question types and supports per-category analysis at scale. This taxonomy lets us ask which selection behavior helps which question type and enables a deployable setting in which a lightweight classifier predicts the category from the question text before routing to a frame-selection preset. In our experiments, we use the seven MLVU multiple-choice task types defined by the evaluation protocol. LongVideoBench~\cite{wu2024longvideobench} provides a complementary long-video QA benchmark; we use its category groups to test whether the fixed relevance--coverage trends also hold outside MLVU\@. Other datasets, such as EgoSchema~\cite{Mangalam_2023_EgoSchema}, probe long-horizon temporal reasoning but do not provide the same convenient category supervision for the type-conditioned routing study.

\subsection{Token reduction via frame selection for VLMs}
A common approach to long-video VLM inference is token reduction: select a small set of frames, or visual tokens, before passing the video to a downstream VLM\@. Adaptive Keyframe Sampling (AKS) combines prompt relevance with keyframe coverage in a plug-and-play selector~\cite{Tang_2025_CVPR}. Flexible Frame Selection learns a policy that focuses on informative frames under context constraints~\cite{Buch_2025_CVPR}. Memory-based systems such as MovieChat compress dense frame streams into persistent memories~\cite{Song_2024_CVPR}, while hierarchical approaches such as VideoTree build structured video representations for reasoning~\cite{Wang_2025_CVPR}. More recent adaptive or agentic sampling systems, such as VideoBrain, learn when and how to acquire additional evidence~\cite{Zou_2026_VideoBrain}. Other query-aware selectors sample from CLIP-like relevance distributions or learn frame rankers from Video-LLM supervision, including Q-Frame~\cite{zhang2025q}, Frame-Voyager~\cite{yu2024frame}, M-LLM-based selectors~\cite{hu2025m}, BOLT~\cite{Liu_2025_CVPR}, and GenS~\cite{yao-etal-2025-generative}. Our work is training-free at the frame-selection stage: it optimizes a submodular relevance--coverage objective and changes the trade-off according to predicted question type.

\subsection{Coverage--diversity objectives and relevance--diversity sampling}
Our objective builds on classic subset-selection formulations that balance relevance with redundancy control. Maximal Marginal Relevance trades off query relevance and novelty~\cite{Carbonell_1998_MMR}, and many coverage or representativeness criteria can be expressed as submodular objectives with efficient greedy approximations under cardinality constraints~\cite{Krause_2014_Submodular}. In long-video QA, this suggests selecting frames that are both question-relevant and representative of the full candidate pool, so that the final subset avoids near-duplicates without ignoring localized evidence.

Recent keyframe samplers instantiate these principles in different ways. AKS emphasizes prompt relevance and coverage under a fixed budget~\cite{Tang_2025_CVPR}, while relevance--diversity methods such as AdaRD-Key use log-determinant or DPP-style criteria to encourage non-redundant selections~\cite{Zhang_2025_AdaRDKey}. These methods demonstrate the importance of balancing query alignment and redundancy control, but they typically apply a single selection rule uniformly across all questions. In contrast, we treat the relevance--coverage balance as question dependent. We define four interpretable presets---Coverage only, Coverage oriented, Relevance only, and Relevance oriented---and study both fixed-preset performance and category-aware routing over them.

%% file: section/3_method.tex
\section{Method}
\label{sec:method}

\subsection{Overview and problem setup}
Given a long video $\mathcal{V}$ and a question $q$, VLM inference is often bottlenecked by the number of input frames and visual tokens. Our goal is to select a subset of $K$ frames that preserves two complementary properties: question relevance and semantic representativeness. Relevance encourages the selected frames to contain answer-specific evidence, while representativeness discourages redundant near-duplicates and encourages coverage of the video's major semantic events.

We formulate frame selection as query-aware subset selection under a cardinality constraint. Specifically, we construct a bounded 1~FPS candidate pool, embed candidates in two complementary spaces for relevance and coverage, and optimize a relevance--coverage set function using greedy maximization. The coverage term is averaged over the candidate pool size so that its magnitude is less sensitive to the number of candidate frames. The resulting objective is monotone submodular, which gives the standard $(1-1/e)$ approximation guarantee for the proposed surrogate objective.

\subsection{Candidate construction and timestamp alignment}
\paragraph{1~FPS candidate seconds.}
For a video with average FPS $f$ and $T$ decoded frames, define duration in seconds as $\lfloor T/f \rfloor$ and construct integer-second candidates:
\[
\mathcal{U} = \{0,1,\dots,\lfloor T/f \rfloor - 1\}.
\]
To bound computation, we cap the candidate count at 1,000. If $|\mathcal{U}|>1{,}000$, we uniformly downsample the second indices to 1,000 (e.g., $\texttt{linspace}(0,|\mathcal{U}|-1,1000)$ followed by integer casting).

\paragraph{Positions vs.\ seconds (alignment invariant).}
Embeddings are stored in the order of the selected seconds. Therefore, the selector operates on \emph{positions}
$i\in\{1,\dots,N\}$ (where $N\le 1{,}000$), which index the embedding set.
A fixed mapping $\pi(i)\mapsto s_i$ converts each position back to its corresponding second index using the \emph{same}
downsampling rule used during embedding extraction.

\paragraph{Loading selected frames.}
Given selected seconds $\{s_i\}$, we load frames at indices $\lfloor s_i\cdot f\rfloor$ (clipped to the valid range),
ensuring consistent timestamp-to-frame alignment between embedding positions and the decoded frames used by the VLM\@.

\subsection{Two embedding spaces}
We precompute two complementary embedding spaces over the $N$ candidates.

\paragraph{Query relevance space (SigLIP).}
For each candidate $i$, we compute a SigLIP~\cite{zhai2023sigmoid} visual embedding $\mathbf{v}_i\in\mathbb{R}^{d_s}$ and for question $q$ a SigLIP text embedding $\mathbf{t}\in\mathbb{R}^{d_s}$. We $\ell_2$-normalize both so $\mathbf{v}_i^\top \mathbf{t}$ is cosine similarity. SigLIP follows the cross-modal contrastive paradigm of CLIP~\cite{radford2021learning} but replaces the softmax contrastive loss with a per-pair sigmoid loss; we use it for relevance scoring because it is well aligned for image-text matching at the single-frame level.

\paragraph{Semantic representativeness space (DINOv2).}
For each candidate $i$, we compute a DINOv2~\cite{oquab2023dinov2} embedding $\mathbf{d}_i\in\mathbb{R}^{d_d}$ and $\ell_2$-normalize it. These embeddings define a semantic similarity space used by a facility-location coverage objective. We use DINOv2 (a self-supervised visual backbone trained without language supervision) for the coverage signal because its features capture intra-visual semantic similarity that is complementary to SigLIP's cross-modal alignment, decoupling ``which frame answers the query'' from ``which frames jointly cover the video.''

\subsection{Objective: relevance plus facility-location coverage}
Let the ground set of candidates be $\mathcal{G}=\{1,\dots,N\}$. We seek a subset $S\subseteq \mathcal{G}$ with $|S|\le K$ maximizing
\begin{equation}
F(S) = \alpha\,R(S) + \beta\,C(S),
\label{eq:final_obj}
\end{equation}
with trade-off weights $\alpha,\beta\ge 0$.

\paragraph{Objective design.}
The objective combines two complementary desiderata for long-video question answering. The relevance term rewards frames that are individually aligned with the question, while the coverage term rewards subsets that represent the semantic structure of the full candidate pool. We average coverage over all candidates because representativeness is measured by how well the selected subset covers the entire video, rather than only by diversity among selected frames; the factor $1/N$ also makes the coverage magnitude comparable across videos with different candidate-pool sizes. The weights \((\alpha,\beta)\) control the trade-off between answer-specific evidence and broad semantic coverage.

\paragraph{Relevance term (modular).}
We define the relevance objective as
\begin{equation}
R(S)=\sum_{i\in S} r_i,
\label{eq:relevance_modular}
\end{equation}
where \(r_i\) is the nonnegative question relevance score of candidate frame \(i\). Specifically, let
\[
a_i = \mathbf{v}_i^\top \mathbf{t}
\]
denote the SigLIP cosine similarity between candidate frame \(i\) and question \(q\), where both embeddings are \(\ell_2\)-normalized. We define
\[
r_i = \max(a_i,0)=\max(\mathbf{v}_i^\top \mathbf{t},0).
\]
This ReLU clipping removes negatively aligned candidates while preserving the ordering among candidates with positive SigLIP similarity. Since \(r_i\ge 0\), adding a frame can only increase or preserve \(R(S)\). Therefore, \(R(S)\) is a nonnegative modular function and is monotone.

\paragraph{Coverage term (facility-location).}
We define the semantic coverage objective as
\begin{equation}
C(S)=\frac{1}{N}\sum_{j\in\mathcal{G}}
\left(
\max\left(b,\max_{i\in S}s_{j,i}\right)-b
\right),
\label{eq:facility_location}
\end{equation}
where \(s_{j,i}\) is the DINOv2 cosine similarity between candidate frames \(j\) and \(i\):
\[
s_{j,i}=\mathbf{d}_j^\top\mathbf{d}_i.
\]
Since the DINOv2 embeddings are \(\ell_2\)-normalized, \(s_{j,i}\in[-1,1]\). We set \(b=-1\), the minimum possible cosine similarity, so that \(C(\emptyset)=0\).

For each candidate \(j\), the inner maximum measures the similarity between \(j\) and its closest selected representative in \(S\). Therefore, \(C(S)\) rewards selected subsets that cover the semantic structure of the full candidate pool. The factor $1/N$ averages this coverage over the candidate set; since $N$ is fixed for a given video, this is a fixed positive rescaling of the standard facility-location objective. Because adding a frame can only increase the maximum similarity for each \(j\), the coverage term is monotone.

\subsection{Greedy maximization with efficient coverage updates}
We optimize \(F(S)\) under the cardinality constraint \(|S|\le K\) using greedy maximization, which is theoretically justified for the monotone submodular objective shown in Sec.~\ref{sec:submod-guarantee}.

Maintain a coverage vector
\[
c_j(S)\triangleq \max\big(b,\max_{i\in S}s_{j,i}\big),
\]
initialized as $c_j(\emptyset)=b$. The marginal coverage gain of adding $i\notin S$ is
\[
\Delta_C(i\mid S)=\frac{1}{N}\sum_{j=1}^{N}\left[\max\!\big(c_j(S), s_{j,i}\big)-c_j(S)\right],
\]
and the total marginal gain is
\[
\Delta(i\mid S)=\alpha\,r_i+\beta\,\Delta_C(i\mid S).
\]
After selecting $i^\star=\arg\max_{i\notin S}\Delta(i\mid S)$, update $c_j\leftarrow \max(c_j,s_{j,i^\star})$ for all $j$.
We return the selected positions sorted in temporal order (equivalently, by their mapped seconds).

\paragraph{Complexity.}
Computing the full DINOv2 similarity matrix costs $O(N^2)$ time and memory.
Given this matrix, a straightforward greedy implementation evaluates all remaining
candidates with an $O(N)$ marginal coverage update at each step, for
$O(KN^2)$ total time. Since we cap the candidate pool at $N\le 1{,}000$, this cost
is practical and easy to vectorize on a GPU\@. Lazy-greedy evaluation~\cite{minoux2005accelerated}
can further reduce computation by reusing cached marginal gains as upper bounds
under submodularity, avoiding unnecessary re-evaluation. Overall, the main
savings come from reducing expensive VLM inference from the full candidate pool
to only $K$ selected frames.

\subsection{Submodularity and greedy approximation guarantee}
\label{sec:submod-guarantee}
We now justify the greedy procedure theoretically.

\paragraph{Monotonicity.}
$R(S)$ is monotone since $r_i\ge 0$. For coverage, for each $j$, the maximum $\max(b,\max_{i\in S}s_{j,i})$ cannot decrease as $S$ grows, and averaging by the fixed positive constant $1/N$ preserves monotonicity. Therefore, for $\alpha,\beta\ge 0$, $F(S)$ is monotone.

\paragraph{Submodularity (diminishing returns).}
$R(S)$ is modular and thus submodular. It suffices to show $C(S)$ is submodular.
For a fixed $j$, define
\[
g_j(S)\triangleq \max\big(b,\max_{i\in S}s_{j,i}\big)-b,
\quad\text{so}\quad
C(S)=\frac{1}{N}\sum_{j\in\mathcal{G}} g_j(S).
\]
Take any $A\subseteq B\subseteq \mathcal{G}$ and $e\notin B$. Let
$m_A=\max(b,\max_{i\in A}s_{j,i})$ and $m_B=\max(b,\max_{i\in B}s_{j,i})$, so $m_A\le m_B$.
Then the marginal gains satisfy
\[
g_j(A\cup\{e\})-g_j(A)=\max(0,s_{j,e}-m_A)\;\ge\;\max(0,s_{j,e}-m_B)=g_j(B\cup\{e\})-g_j(B),
\]
which proves each $g_j$ is submodular. Since $C(S)$ is a nonnegative linear combination of the $g_j$ terms with fixed coefficient $1/N$, $C(S)$ is submodular. A nonnegative linear combination preserves submodularity, so $F(S)$ is monotone submodular.

\paragraph{Greedy guarantee.}
The averaged coverage term satisfies $C(\emptyset)=0$, so $F(\emptyset)=0$. Maximizing this normalized monotone submodular $F$ under $|S|\le K$ with greedy yields the classic bound
\begin{equation}
F(S_{\text{greedy}})\;\ge\;\left(1-\frac{1}{e}\right)F(S^\star),
\label{eq:greedy_bound}
\end{equation}
where $S^\star$ is the optimal subset of size at most $K$.

\paragraph{Match to implementation.}
Our maintained coverage $c_j$ equals $c_j(S)=\max\big(b,\max_{i\in S}s_{j,i}\big)$, and the per-step score
\[
\Delta(e\mid S)=\alpha\,r_e+\beta\frac{1}{N}\sum_{j\in\mathcal{G}}\left[\max\!\big(c_j(S),s_{j,e}\big)-c_j(S)\right]
\]
is exactly the marginal gain $F(S\cup\{e\})-F(S)$, so Alg.~\ref{alg:greedy} is the greedy algorithm for Eq.~\eqref{eq:final_obj}.

\subsection{Question-type adaptive selection}
\label{sec:qtype-method}

Different question types may prefer different relevance--coverage trade-offs: summary or topic questions often require broad semantic coverage, whereas needle-style questions may benefit from highly question-relevant frames. We therefore evaluate both fixed presets and category-aware routing policies.

\paragraph{Fixed strategy set.}
All methods select $K$ frames from the same candidate pool. In addition to Uniform sampling and AKS~\cite{Tang_2025_CVPR} as reference baselines, we instantiate four greedy presets from Eq.~\eqref{eq:final_obj}:
\[
\begin{array}{ll}
\text{Coverage only}: (\alpha,\beta)=(0,1), &
\text{Coverage oriented}: (\alpha,\beta)=(0.5,1),\\
\text{Relevance only}: (\alpha,\beta)=(1,0), &
\text{Relevance oriented}: (\alpha,\beta)=(1,0.5).
\end{array}
\]
We denote these four greedy presets by $\mathcal{S}$. Uniform and AKS are reported as baselines, but are not included in $\mathcal{S}$ when computing oracle or adaptive routing results.

\paragraph{Oracle upper bound.}
To estimate the headroom from perfect category-aware strategy selection, we report an Oracle Strategy. For each frame budget $K$ and ground-truth category $c$, it retrospectively chooses the best-performing preset:
\[
s^{\star}_{\mathrm{oracle}}(K,c)
=
\arg\max_{s\in\mathcal{S}}
\mathrm{Acc}_{\mathrm{eval}}(K,s\mid c).
\]
Because this uses ground-truth categories and evaluation-set outcomes, it is not deployable; we use it only as an upper-bound diagnostic.

\paragraph{Deployable MLVU routing.}
For MLVU, we also evaluate a deployable adaptive policy on a 40/20/40 train/validation/test split. A lightweight text-only classifier is trained on the train split to predict the question category $\hat c$. On the validation split, we route questions by predicted category and select one budget-agnostic preset per predicted category:
\[
s^{\star}_{\mathrm{val}}(\hat c)
=
\arg\max_{s\in\mathcal{S}}
\bar{\mathrm{Acc}}_{\mathrm{val}}(s\mid \hat c),
\]
where $\bar{\mathrm{Acc}}_{\mathrm{val}}$ averages over the evaluated frame budgets. After validation, both the classifier and the category-to-strategy mapping are fixed. At test time, each question is routed only by its predicted category and uses the validation-selected preset for every budget $K$. This policy is therefore strictly more constrained than Oracle Strategy, which selects per category and per budget using evaluation outcomes.

%% file: section/4_exp.tex
\section{Experiments}
\label{sec:experiments}

\subsection{Questions and evaluation protocol}
\label{sec:exp_setup}

\begin{figure}[t]
    \centering

    \begin{subfigure}[t]{0.49\linewidth}
        \centering
        \includegraphics[width=\linewidth]{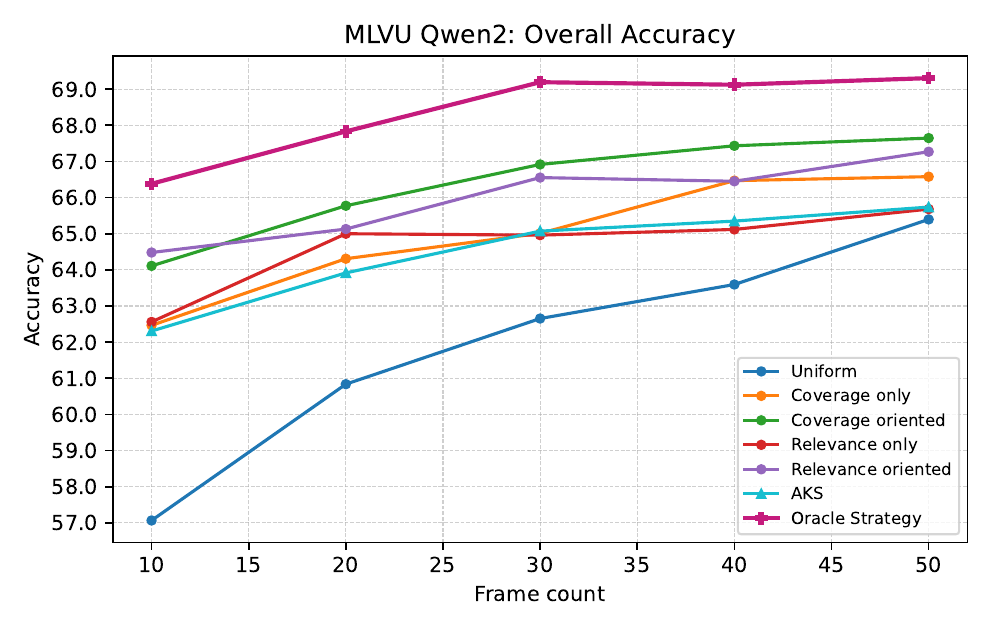}
        \caption{MLVU, Qwen2-VL.}
        \label{fig:main_budget_mlvu_qwen2}
    \end{subfigure}
    \hfill
    \begin{subfigure}[t]{0.49\linewidth}
        \centering
        \includegraphics[width=\linewidth]{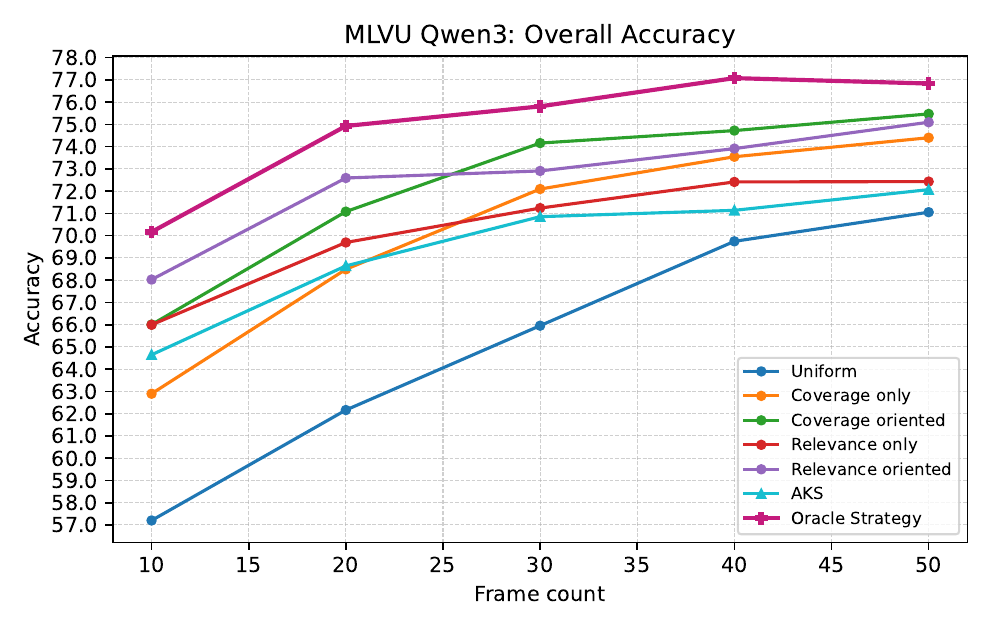}
        \caption{MLVU, Qwen3-VL.}
        \label{fig:main_budget_mlvu_qwen3}
    \end{subfigure}

    \vspace{0.6em}

    \begin{subfigure}[t]{0.49\linewidth}
        \centering
        \includegraphics[width=\linewidth]{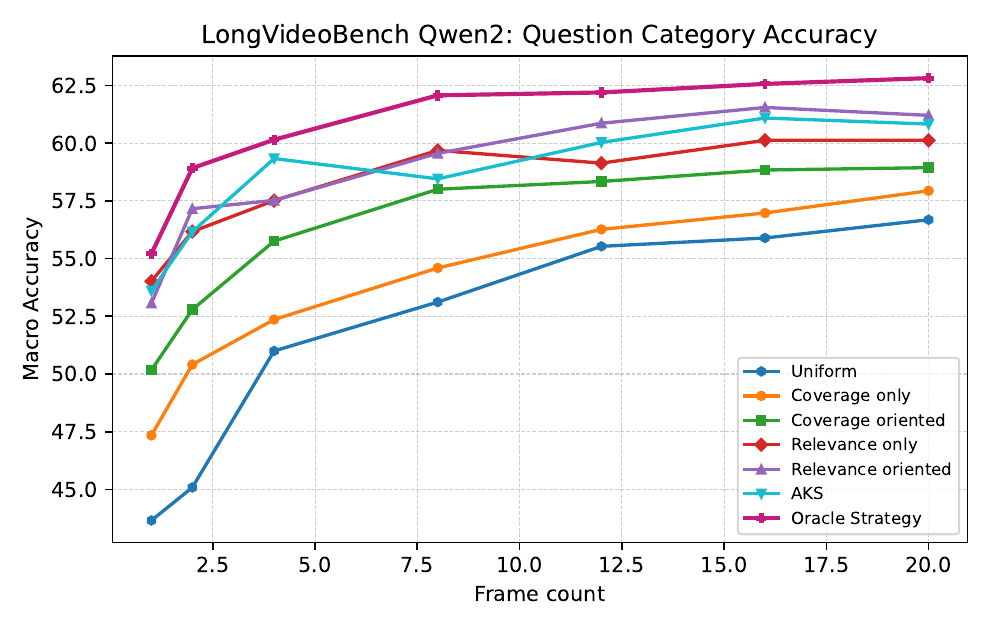}
        \caption{LongVideoBench, Qwen2-VL.}
        \label{fig:main_budget_lvb_qwen2}
    \end{subfigure}
    \hfill
    \begin{subfigure}[t]{0.49\linewidth}
        \centering
        \includegraphics[width=\linewidth]{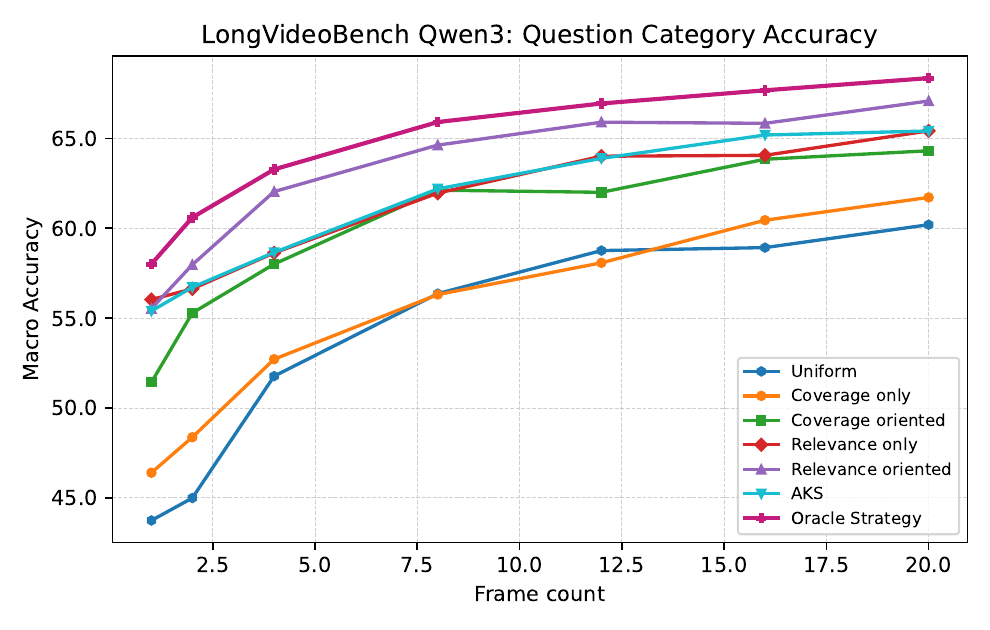}
        \caption{LongVideoBench, Qwen3-VL.}
        \label{fig:main_budget_lvb_qwen3}
    \end{subfigure}

    \caption{\textbf{Accuracy as a function of the selected frame budget.}
    All methods select $K$ frames from the same 1\,FPS candidate pool. MLVU panels report aggregate multiple-choice accuracy, while LongVideoBench panels report macro accuracy over category groups. Uniform and AKS are reference baselines; the four relevance--coverage presets are instances of our greedy submodular selector with different $(\alpha,\beta)$ trade-offs. Oracle Strategy is a post-hoc category-aware upper bound over the four greedy presets and is not ranked as a deployable method.}
    \label{fig:main_budget_all}
\end{figure}

\begin{table}[t]
\centering
\caption{\textbf{Aggregate accuracy at representative frame budgets.}
MLVU reports aggregate multiple-choice accuracy; LongVideoBench reports macro accuracy over category groups. Oracle Strategy is a post-hoc upper bound that chooses the best greedy preset for each category and budget using evaluation outcomes, and is excluded from best/second-best marking. Bold and underline indicate the best and second-best non-oracle rows among Uniform, AKS, and the four greedy presets.}

\label{tab:main_results}
\small
\setlength{\tabcolsep}{3pt}
\resizebox{\linewidth}{!}{
\begin{tabular}{l ccc ccc ccc ccc}
\toprule
 & \multicolumn{6}{c}{MLVU} & \multicolumn{6}{c}{LongVideoBench} \\
\cmidrule(lr){2-7} \cmidrule(lr){8-13}
 & \multicolumn{3}{c}{Qwen2-VL} & \multicolumn{3}{c}{Qwen3-VL} & \multicolumn{3}{c}{Qwen2-VL} & \multicolumn{3}{c}{Qwen3-VL} \\
\cmidrule(lr){2-4} \cmidrule(lr){5-7} \cmidrule(lr){8-10} \cmidrule(lr){11-13}
Method & $K{=}10$ & $K{=}30$ & $K{=}50$ & $K{=}10$ & $K{=}30$ & $K{=}50$ & $K{=}4$ & $K{=}12$ & $K{=}20$ & $K{=}4$ & $K{=}12$ & $K{=}20$ \\
\midrule
Uniform & 57.07 & 62.65 & 65.40 & 57.20 & 65.95 & 71.05 & 50.57 & 55.14 & 56.22 & 50.80 & 57.69 & 59.08 \\
AKS~\cite{Tang_2025_CVPR} & 62.31 & 65.07 & 65.74 & 64.65 & 70.85 & 72.07 & \textbf{58.75} & \underline{59.43} & \underline{59.99} & 58.00 & \underline{63.92} & \underline{65.02} \\
\midrule
Coverage only & 62.46 & 65.01 & 66.58 & 62.90 & 72.10 & 74.40 & 52.00 & 55.84 & 57.50 & 52.01 & 57.57 & 60.89 \\
Coverage oriented & \underline{64.11} & \textbf{66.92} & \textbf{67.65} & \underline{66.00} & \textbf{74.16} & \textbf{75.47} & 55.36 & 57.67 & 58.39 & 57.49 & 61.23 & 63.82 \\
Relevance only & 62.56 & 64.96 & 65.68 & 65.99 & 71.24 & 72.43 & \underline{57.47} & 58.55 & 59.80 & \underline{58.60} & 63.73 & 64.92 \\
Relevance oriented & \textbf{64.48} & \underline{66.55} & \underline{67.27} & \textbf{68.03} & \underline{72.91} & \underline{75.10} & 56.95 & \textbf{60.49} & \textbf{60.92} & \textbf{62.01} & \textbf{65.23} & \textbf{66.39} \\
\midrule
Oracle Strategy & 66.38 & 69.19 & 69.30 & 70.17 & 75.81 & 76.84 & 59.25 & 61.37 & 61.51 & 62.74 & 66.71 & 66.78 \\
\bottomrule
\end{tabular}}
\end{table}

We evaluate whether long-video frame selection should use a single fixed rule or adapt its relevance--coverage trade-off to the question type. Our experiments address three questions: (i) whether relevance--coverage selection improves QA under fixed frame budgets, (ii) whether relevance and facility-location coverage provide complementary signals across datasets and VLM backbones, and (iii) whether a deployable question-type router can recover part of the headroom suggested by a category oracle.

We evaluate Qwen2-VL~\cite{wang2024qwen2} and Qwen3-VL~\cite{bai2025qwen3} on MLVU and LongVideoBench. For MLVU, we follow the benchmark taxonomy and use seven multiple-choice task types: \texttt{plotQA}, \texttt{needle}, \texttt{ego}, \texttt{count}, \texttt{order}, \texttt{anomaly\_reco}, and \texttt{topic\_reasoning}. For LongVideoBench, we report macro accuracy over the benchmark category groups. The held-out MLVU routing experiment in Table~\ref{tab:deployable_results} reports category-average accuracy.

All methods select exactly $K$ frames from the same 1\,FPS candidate pool, capped at 1,000 frames per video, and use the same VLM prompt and decoding protocol. We compare Uniform and AKS~\cite{Tang_2025_CVPR} with the four greedy presets from Sec.~\ref{sec:qtype-method}: Coverage only, Coverage oriented, Relevance only, and Relevance oriented. Oracle Strategy retrospectively selects the best greedy preset for each category and budget using evaluation outcomes, and is reported only as a non-deployable upper bound. For MLVU, we additionally evaluate a deployable adaptive strategy whose category-to-strategy routing policy is selected on validation data and fixed before held-out test evaluation.

\subsection{Main results: fixed relevance--coverage selection}
\label{sec:main_results}

Fig.~\ref{fig:main_budget_all} and Table~\ref{tab:main_results} show that frame selection is most important when the VLM can observe only a small number of frames. At the tightest representative budget, the strongest proposed greedy preset improves over Uniform by $+7.41\%$ and $+10.83\%$ on MLVU with Qwen2-VL and Qwen3-VL, and by $+6.90\%$ and $+11.21\%$ on LongVideoBench with Qwen2-VL and Qwen3-VL, respectively. These gains are larger than the remaining gains from simply increasing the frame budget for several baselines, indicating that \emph{which} frames are selected can matter as much as \emph{how many} frames are used.

On MLVU, the best greedy preset exceeds AKS by an average of $+1.98$\% for Qwen2-VL and $+3.36$\% for Qwen3-VL across the three representative budgets in Table~\ref{tab:main_results}. On LongVideoBench, the best greedy preset exceeds AKS by $+2.23$\% on average with Qwen3-VL and is competitive with AKS on Qwen2-VL; the main exception is the very small-budget LongVideoBench/Qwen2 setting ($K{=}4$), where AKS remains the best non-oracle method. This exception is useful: it shows that the result is not driven by a uniformly dominant preset, but by matching the selector to the question and budget.

The table further supports the method design in Sec.~\ref{sec:method}. Pure relevance and pure coverage are both useful, but neither is consistently optimal. Among the four greedy presets, Relevance oriented is the best preset in 7 of the 12 representative settings, Coverage oriented is best in 4 settings, and Relevance only is best in 1 setting. This pattern suggests that answer-specific alignment and semantic representativeness provide complementary signals. In practice, the best fixed selectors usually retain some weight on the other signal rather than relying exclusively on relevance or coverage.

\subsection{Oracle Strategy as a headroom diagnostic}
\label{sec:oracle_strategy}

Oracle Strategy is included only to measure the value of category-aware strategy selection. It uses ground-truth category labels and evaluation-set outcomes to choose the best greedy preset for each category and frame budget, and therefore should not be interpreted as a test-time method.

The oracle gap is nevertheless informative. Across the 12 representative settings in Table~\ref{tab:main_results}, Oracle Strategy exceeds the best single greedy preset by $0.39\%$--$2.27\%$, with an average gap of $1.40\%$. The gap shows that no single relevance--coverage trade-off is uniformly preferred by all question categories. Instead, categories differ in whether they benefit more from localized question relevance or broad temporal/semantic coverage. This is precisely the situation targeted by the adaptive routing policy evaluated next.

\subsection{Deployable adaptive strategy on the MLVU held-out test split}
\label{sec:deployable_results}

The full-dataset oracle analysis answers how much headroom exists under perfect category-aware routing. We next test whether a realistic routing policy can recover part of this headroom. For MLVU, we split the data into train/validation/test partitions. A lightweight text-only classifier is trained on the train split to predict the question type. The validation split is then used to choose one greedy preset per predicted category by averaging validation accuracy across frame budgets. After validation, the classifier and category-to-strategy mapping are fixed; test questions are routed only by their predicted category, and the same routing policy is reused for every test budget.

\begin{figure}[t]
    \centering
    \begin{subfigure}[t]{0.49\linewidth}
        \centering
        \includegraphics[width=\linewidth]{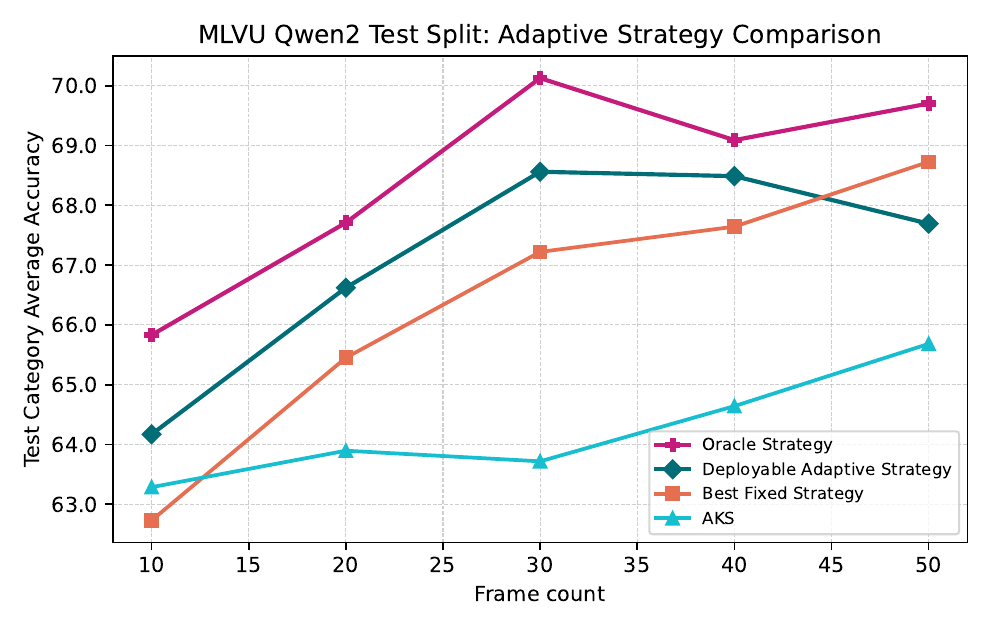}
        \caption{Qwen2-VL.}
        \label{fig:mlvu_test_split_adaptive_qwen2}
    \end{subfigure}
    \hfill
    \begin{subfigure}[t]{0.49\linewidth}
        \centering
        \includegraphics[width=\linewidth]{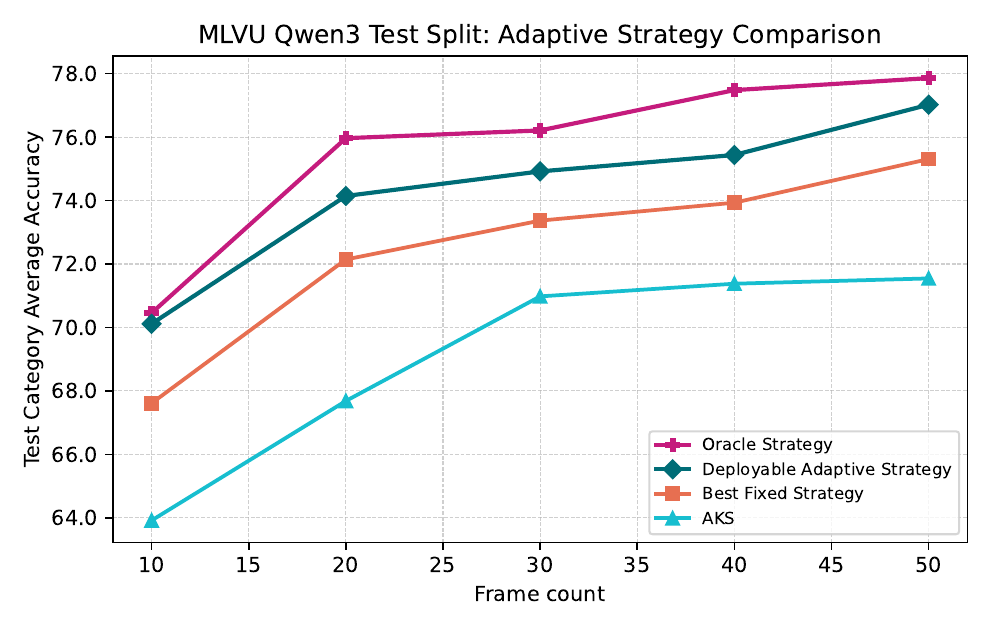}
        \caption{Qwen3-VL.}
        \label{fig:mlvu_test_split_adaptive_qwen3}
    \end{subfigure}
    \caption{\textbf{Held-out MLVU test evaluation of deployable adaptive routing.}
    Deployable Adaptive Strategy uses a text-only question-type classifier and a validation-selected category-to-strategy mapping. Best Fixed is the strongest single fixed preset under the same held-out averaging protocol. Oracle Strategy uses ground-truth test categories and test-set outcomes to choose the best strategy for each category and budget, and is therefore a non-deployable upper bound.}
    \label{fig:mlvu_test_split_adaptive}
\end{figure}

\paragraph{Headroom realized vs.\ headroom available.}
For MLVU, we report both the oracle upper bound and the deployable adaptive result. To summarize how much of the oracle headroom is realized by practical routing beyond the strongest fixed strategy, we compute
\[
\rho_{\mathrm{fixed}}=
\frac{\mathrm{Acc}_{\mathrm{deploy}}-\mathrm{Acc}_{\mathrm{best\ fixed}}}
{\mathrm{Acc}_{\mathrm{oracle}}-\mathrm{Acc}_{\mathrm{best\ fixed}}}.
\]
Here, $\mathrm{Acc}_{\mathrm{best\ fixed}}$ is the best single fixed strategy under the same held-out test-split averaging protocol. Deployable Adaptive Strategy improves over Best Fixed by $+0.76$\% with Qwen2-VL and $+1.86$\% with Qwen3-VL, realizing $35.5\%$ and $59.6\%$ of the oracle-over-best-fixed headroom, respectively. Relative to AKS, it improves by $+2.87\%$ and $+5.23\%$, showing that category-aware routing provides gains beyond both AKS and the strongest fixed strategy.

\begin{table}[t]
\centering
\caption{\textbf{Held-out MLVU test-split comparison.}
Results are category-average accuracy averaged over $K\in\{10,20,30,40,50\}$. Best Fixed is the strongest single fixed strategy under the same protocol. Deployable Adaptive routes questions by predicted category to a validation-selected strategy, whereas Oracle uses ground-truth categories and test-set outcomes and is therefore a non-deployable upper bound. Oracle is excluded from ranking; bold and underline denote the best and second-best non-oracle methods.}
\label{tab:deployable_results}
\small
\begin{tabular}{lcccc}
\toprule
Backbone & AKS & Best Fixed & Deployable Adaptive & Oracle \\
\midrule
Qwen2-VL & 64.24\,\% & \underline{66.35\,\%} & \textbf{67.11\,\%} & 68.49\,\% \\
Qwen3-VL & 69.10\,\% & \underline{72.47\,\%} & \textbf{74.33\,\%} & 75.59\,\% \\
\bottomrule
\end{tabular}
\end{table}

\paragraph{Takeaway.}
Long-video frame selection should not be treated as a one-size-fits-all preprocessing step. The proposed relevance--coverage objective improves fixed-budget frame selection, the best relevance--coverage trade-off varies across settings and categories, and a validation-selected question-type router recovers a meaningful portion of the oracle headroom on a held-out test split. The practical conclusion is that adaptive routing over a small set of interpretable submodular selectors provides a stronger and more deployable alternative to uniform sampling or a single fixed keyframe strategy.

%% file: section/5_conclusion.tex
\section{Conclusion}
We presented a question-adaptive greedy frame-selection method for long-video VLM inference. The selector combines SigLIP-based query relevance with DINOv2-based facility-location coverage, producing a simple monotone submodular surrogate that can be optimized efficiently by greedy maximization under a fixed frame budget.
The main empirical finding is that the relevance--coverage trade-off should not be fixed globally. Across MLVU and LongVideoBench, mixed relevance--coverage presets improve low-budget frame selection over uniform sampling and are competitive with or stronger than AKS\@. On the MLVU held-out split, a validation-selected router over predicted question types further improves over the best fixed strategy and recovers part of the oracle headroom, showing that practical question-adaptive routing is feasible without using test-set outcomes.
\paragraph{Broader impacts.}
More efficient frame selection can reduce the compute cost of long-video VLM inference and make long-video QA more accessible. At the same time, improvements in long-video understanding could be misused in surveillance or other sensitive settings; responsible deployment should respect dataset/model licenses, privacy constraints, and the safety policies of the underlying VLMs.
\paragraph{Limitations and future work.}
The deployable adaptive-routing experiment relies on MLVU question-type supervision; extending the same end-to-end routing protocol to benchmarks without explicit task labels will require either learned latent question groups or externally defined taxonomies. Oracle Strategy is intentionally non-deployable and should be interpreted only as a diagnostic upper bound. The current strategy set is small and interpretable, but richer routers could condition on uncertainty, video length, or iterative evidence acquisition. Finally, our implementation uses a full DINOv2 similarity matrix over a candidate pool capped at 1,000 frames; larger candidate pools would benefit from approximate or lazy-greedy variants. Broader evaluation on additional long-video benchmarks such as Video-MME and EgoSchema is an important next step.

%% file: checklist.tex
\section*{NeurIPS Paper Checklist}

\begin{enumerate}

\item {\bf Claims}
    \item[] Question: Do the main claims made in the abstract and introduction accurately reflect the paper's contributions and scope?
    \item[] Answer: \answerYes{}
    \item[] Justification: The abstract and introduction state the relevance--coverage objective, the distinction between oracle and deployable routing, and the experimental scope on MLVU and LongVideoBench.
    \item[] Guidelines:
    \begin{itemize}
        \item The answer \answerNA{} means that the abstract and introduction do not include the claims made in the paper.
        \item The abstract and/or introduction should clearly state the claims made, including the contributions made in the paper and important assumptions and limitations. A \answerNo{} or \answerNA{} answer to this question will not be perceived well by the reviewers. 
        \item The claims made should match theoretical and experimental results, and reflect how much the results can be expected to generalize to other settings. 
        \item It is fine to include aspirational goals as motivation as long as it is clear that these goals are not attained by the paper. 
    \end{itemize}

\item {\bf Limitations}
    \item[] Question: Does the paper discuss the limitations of the work performed by the authors?
    \item[] Answer: \answerYes{}
    \item[] Justification: The conclusion includes limitations on deployable routing supervision, the non-deployable oracle, the small preset set, and the quadratic DINOv2 similarity computation.
    \item[] Guidelines:
    \begin{itemize}
        \item The answer \answerNA{} means that the paper has no limitation while the answer \answerNo{} means that the paper has limitations, but those are not discussed in the paper. 
        \item The authors are encouraged to create a separate ``Limitations'' section in their paper.
        \item The paper should point out any strong assumptions and how robust the results are to violations of these assumptions (e.g., independence assumptions, noiseless settings, model well-specification, asymptotic approximations only holding locally). The authors should reflect on how these assumptions might be violated in practice and what the implications would be.
        \item The authors should reflect on the scope of the claims made, e.g., if the approach was only tested on a few datasets or with a few runs. In general, empirical results often depend on implicit assumptions, which should be articulated.
        \item The authors should reflect on the factors that influence the performance of the approach. For example, a facial recognition algorithm may perform poorly when image resolution is low or images are taken in low lighting. Or a speech-to-text system might not be used reliably to provide closed captions for online lectures because it fails to handle technical jargon.
        \item The authors should discuss the computational efficiency of the proposed algorithms and how they scale with dataset size.
        \item If applicable, the authors should discuss possible limitations of their approach to address problems of privacy and fairness.
        \item While the authors might fear that complete honesty about limitations might be used by reviewers as grounds for rejection, a worse outcome might be that reviewers discover limitations that aren't acknowledged in the paper. The authors should use their best judgment and recognize that individual actions in favor of transparency play an important role in developing norms that preserve the integrity of the community. Reviewers will be specifically instructed to not penalize honesty concerning limitations.
    \end{itemize}

\item {\bf Theory assumptions and proofs}
    \item[] Question: For each theoretical result, does the paper provide the full set of assumptions and a complete (and correct) proof?
    \item[] Answer: \answerYes{}
    \item[] Justification: Sec.~\ref{sec:method} states the assumptions $\alpha,\beta\ge0$, the normalized facility-location objective, and the proof of monotonicity, submodularity, and the greedy approximation guarantee.
    \item[] Guidelines:
    \begin{itemize}
        \item The answer \answerNA{} means that the paper does not include theoretical results. 
        \item All the theorems, formulas, and proofs in the paper should be numbered and cross-referenced.
        \item All assumptions should be clearly stated or referenced in the statement of any theorems.
        \item The proofs can either appear in the main paper or the supplemental material, but if they appear in the supplemental material, the authors are encouraged to provide a short proof sketch to provide intuition. 
        \item Inversely, any informal proof provided in the core of the paper should be complemented by formal proofs provided in appendix or supplemental material.
        \item Theorems and Lemmas that the proof relies upon should be properly referenced. 
    \end{itemize}

    \item {\bf Experimental result reproducibility}
    \item[] Question: Does the paper fully disclose all the information needed to reproduce the main experimental results of the paper to the extent that it affects the main claims and/or conclusions of the paper (regardless of whether the code and data are provided or not)?
    \item[] Answer: \answerYes{}
    \item[] Justification: Secs.~\ref{sec:method} and~\ref{sec:experiments} specify candidate construction, embeddings, objective weights, frame budgets, baselines, held-out splits, and routing protocol for the reported results.
    \item[] Guidelines:
    \begin{itemize}
        \item The answer \answerNA{} means that the paper does not include experiments.
        \item If the paper includes experiments, a \answerNo{} answer to this question will not be perceived well by the reviewers: Making the paper reproducible is important, regardless of whether the code and data are provided or not.
        \item If the contribution is a dataset and\slash or model, the authors should describe the steps taken to make their results reproducible or verifiable. 
        \item Depending on the contribution, reproducibility can be accomplished in various ways. For example, if the contribution is a novel architecture, describing the architecture fully might suffice, or if the contribution is a specific model and empirical evaluation, it may be necessary to either make it possible for others to replicate the model with the same dataset, or provide access to the model. In general. releasing code and data is often one good way to accomplish this, but reproducibility can also be provided via detailed instructions for how to replicate the results, access to a hosted model (e.g., in the case of a large language model), releasing of a model checkpoint, or other means that are appropriate to the research performed.
        \item While NeurIPS does not require releasing code, the conference does require all submissions to provide some reasonable avenue for reproducibility, which may depend on the nature of the contribution. For example
        \begin{enumerate}
            \item If the contribution is primarily a new algorithm, the paper should make it clear how to reproduce that algorithm.
            \item If the contribution is primarily a new model architecture, the paper should describe the architecture clearly and fully.
            \item If the contribution is a new model (e.g., a large language model), then there should either be a way to access this model for reproducing the results or a way to reproduce the model (e.g., with an open-source dataset or instructions for how to construct the dataset).
            \item We recognize that reproducibility may be tricky in some cases, in which case authors are welcome to describe the particular way they provide for reproducibility. In the case of closed-source models, it may be that access to the model is limited in some way (e.g., to registered users), but it should be possible for other researchers to have some path to reproducing or verifying the results.
        \end{enumerate}
    \end{itemize}

\item {\bf Open access to data and code}
    \item[] Question: Does the paper provide open access to the data and code, with sufficient instructions to faithfully reproduce the main experimental results, as described in supplemental material?
    \item[] Answer: \answerNo{}
    \item[] Justification: This draft does not yet include an anonymized code release or reproduction repository. The experiments use public benchmarks and publicly described models, but code-release instructions should be added if code will be shared.
    \item[] Guidelines:
    \begin{itemize}
        \item The answer \answerNA{} means that paper does not include experiments requiring code.
        \item Please see the NeurIPS code and data submission guidelines (\url{https://neurips.cc/public/guides/CodeSubmissionPolicy}) for more details.
        \item While we encourage the release of code and data, we understand that this might not be possible, so \answerNo{} is an acceptable answer. Papers cannot be rejected simply for not including code, unless this is central to the contribution (e.g., for a new open-source benchmark).
        \item The instructions should contain the exact command and environment needed to run to reproduce the results. See the NeurIPS code and data submission guidelines (\url{https://neurips.cc/public/guides/CodeSubmissionPolicy}) for more details.
        \item The authors should provide instructions on data access and preparation, including how to access the raw data, preprocessed data, intermediate data, and generated data, etc.
        \item The authors should provide scripts to reproduce all experimental results for the new proposed method and baselines. If only a subset of experiments are reproducible, they should state which ones are omitted from the script and why.
        \item At submission time, to preserve anonymity, the authors should release anonymized versions (if applicable).
        \item Providing as much information as possible in supplemental material (appended to the paper) is recommended, but including URLs to data and code is permitted.
    \end{itemize}

\item {\bf Experimental setting/details}
    \item[] Question: Does the paper specify all the training and test details (e.g., data splits, hyperparameters, how they were chosen, type of optimizer) necessary to understand the results?
    \item[] Answer: \answerYes{}
    \item[] Justification: The experimental setup specifies datasets, frame budgets, candidate-pool construction, backbones, baselines, the 40/20/40 MLVU split, and the validation-selected routing policy.
    \item[] Guidelines:
    \begin{itemize}
        \item The answer \answerNA{} means that the paper does not include experiments.
        \item The experimental setting should be presented in the core of the paper to a level of detail that is necessary to appreciate the results and make sense of them.
        \item The full details can be provided either with the code, in appendix, or as supplemental material.
    \end{itemize}

\item {\bf Experiment statistical significance}
    \item[] Question: Does the paper report error bars suitably and correctly defined or other appropriate information about the statistical significance of the experiments?
    \item[] Answer: \answerNo{}
    \item[] Justification: The paper reports deterministic evaluation results without error bars or confidence intervals. This should be justified by the high cost of repeated VLM inference, or supplemented with bootstrap intervals if feasible.
    \item[] Guidelines:
    \begin{itemize}
        \item The answer \answerNA{} means that the paper does not include experiments.
        \item The authors should answer \answerYes{} if the results are accompanied by error bars, confidence intervals, or statistical significance tests, at least for the experiments that support the main claims of the paper.
        \item The factors of variability that the error bars are capturing should be clearly stated (for example, train/test split, initialization, random drawing of some parameter, or overall run with given experimental conditions).
        \item The method for calculating the error bars should be explained (closed form formula, call to a library function, bootstrap, etc.)
        \item The assumptions made should be given (e.g., Normally distributed errors).
        \item It should be clear whether the error bar is the standard deviation or the standard error of the mean.
        \item It is OK to report 1-sigma error bars, but one should state it. The authors should preferably report a 2-sigma error bar than state that they have a 96\% CI, if the hypothesis of Normality of errors is not verified.
        \item For asymmetric distributions, the authors should be careful not to show in tables or figures symmetric error bars that would yield results that are out of range (e.g., negative error rates).
        \item If error bars are reported in tables or plots, the authors should explain in the text how they were calculated and reference the corresponding figures or tables in the text.
    \end{itemize}

\item {\bf Experiments compute resources}
    \item[] Question: For each experiment, does the paper provide sufficient information on the computer resources (type of compute workers, memory, time of execution) needed to reproduce the experiments?
    \item[] Answer: \answerNo{}
    \item[] Justification: The method section reports algorithmic complexity, but this draft does not yet provide hardware, memory, wall-clock, or total compute details for the experiments.
    \item[] Guidelines:
    \begin{itemize}
        \item The answer \answerNA{} means that the paper does not include experiments.
        \item The paper should indicate the type of compute workers CPU or GPU, internal cluster, or cloud provider, including relevant memory and storage.
        \item The paper should provide the amount of compute required for each of the individual experimental runs as well as estimate the total compute. 
        \item The paper should disclose whether the full research project required more compute than the experiments reported in the paper (e.g., preliminary or failed experiments that didn't make it into the paper). 
    \end{itemize}
    
\item {\bf Code of ethics}
    \item[] Question: Does the research conducted in the paper conform, in every respect, with the NeurIPS Code of Ethics \url{https://neurips.cc/public/EthicsGuidelines}?
    \item[] Answer: \answerYes{}
    \item[] Justification: The work uses public benchmarks and publicly available VLM backbones for evaluation, does not collect new human-subject data, and is consistent with the stated research-use setting.
    \item[] Guidelines:
    \begin{itemize}
        \item The answer \answerNA{} means that the authors have not reviewed the NeurIPS Code of Ethics.
        \item If the authors answer \answerNo, they should explain the special circumstances that require a deviation from the Code of Ethics.
        \item The authors should make sure to preserve anonymity (e.g., if there is a special consideration due to laws or regulations in their jurisdiction).
    \end{itemize}

\item {\bf Broader impacts}
    \item[] Question: Does the paper discuss both potential positive societal impacts and negative societal impacts of the work performed?
    \item[] Answer: \answerYes{}
    \item[] Justification: The conclusion discusses positive impacts from reduced inference cost as well as possible misuse risks inherited from improved long-video understanding, including surveillance-sensitive settings.
    \item[] Guidelines:
    \begin{itemize}
        \item The answer \answerNA{} means that there is no societal impact of the work performed.
        \item If the authors answer \answerNA{} or \answerNo, they should explain why their work has no societal impact or why the paper does not address societal impact.
        \item Examples of negative societal impacts include potential malicious or unintended uses (e.g., disinformation, generating fake profiles, surveillance), fairness considerations (e.g., deployment of technologies that could make decisions that unfairly impact specific groups), privacy considerations, and security considerations.
        \item The conference expects that many papers will be foundational research and not tied to particular applications, let alone deployments. However, if there is a direct path to any negative applications, the authors should point it out. For example, it is legitimate to point out that an improvement in the quality of generative models could be used to generate Deepfakes for disinformation. On the other hand, it is not needed to point out that a generic algorithm for optimizing neural networks could enable people to train models that generate Deepfakes faster.
        \item The authors should consider possible harms that could arise when the technology is being used as intended and functioning correctly, harms that could arise when the technology is being used as intended but gives incorrect results, and harms following from (intentional or unintentional) misuse of the technology.
        \item If there are negative societal impacts, the authors could also discuss possible mitigation strategies (e.g., gated release of models, providing defenses in addition to attacks, mechanisms for monitoring misuse, mechanisms to monitor how a system learns from feedback over time, improving the efficiency and accessibility of ML).
    \end{itemize}
    
\item {\bf Safeguards}
    \item[] Question: Does the paper describe safeguards that have been put in place for responsible release of data or models that have a high risk for misuse (e.g., pre-trained language models, image generators, or scraped datasets)?
    \item[] Answer: \answerNA{}
    \item[] Justification: The paper does not release a high-risk pretrained model, scraped dataset, or generative system; it evaluates a frame-selection method on existing benchmarks.
    \item[] Guidelines:
    \begin{itemize}
        \item The answer \answerNA{} means that the paper poses no such risks.
        \item Released models that have a high risk for misuse or dual-use should be released with necessary safeguards to allow for controlled use of the model, for example by requiring that users adhere to usage guidelines or restrictions to access the model or implementing safety filters. 
        \item Datasets that have been scraped from the Internet could pose safety risks. The authors should describe how they avoided releasing unsafe images.
        \item We recognize that providing effective safeguards is challenging, and many papers do not require this, but we encourage authors to take this into account and make a best faith effort.
    \end{itemize}

\item {\bf Licenses for existing assets}
    \item[] Question: Are the creators or original owners of assets (e.g., code, data, models), used in the paper, properly credited and are the license and terms of use explicitly mentioned and properly respected?
    \item[] Answer: \answerNo{}
    \item[] Justification: The draft cites the datasets, models, and baselines, but it does not yet explicitly list licenses or terms of use for all existing assets. Add an asset/license paragraph before submission if required.
    \item[] Guidelines:
    \begin{itemize}
        \item The answer \answerNA{} means that the paper does not use existing assets.
        \item The authors should cite the original paper that produced the code package or dataset.
        \item The authors should state which version of the asset is used and, if possible, include a URL.
        \item The name of the license (e.g., CC-BY 4.0) should be included for each asset.
        \item For scraped data from a particular source (e.g., website), the copyright and terms of service of that source should be provided.
        \item If assets are released, the license, copyright information, and terms of use in the package should be provided. For popular datasets, \url{paperswithcode.com/datasets} has curated licenses for some datasets. Their licensing guide can help determine the license of a dataset.
        \item For existing datasets that are re-packaged, both the original license and the license of the derived asset (if it has changed) should be provided.
        \item If this information is not available online, the authors are encouraged to reach out to the asset's creators.
    \end{itemize}

\item {\bf New assets}
    \item[] Question: Are new assets introduced in the paper well documented and is the documentation provided alongside the assets?
    \item[] Answer: \answerNA{}
    \item[] Justification: The submission does not introduce or release a new dataset, model checkpoint, or other standalone asset beyond the proposed method description.
    \item[] Guidelines:
    \begin{itemize}
        \item The answer \answerNA{} means that the paper does not release new assets.
        \item Researchers should communicate the details of the dataset\slash code\slash model as part of their submissions via structured templates. This includes details about training, license, limitations, etc. 
        \item The paper should discuss whether and how consent was obtained from people whose asset is used.
        \item At submission time, remember to anonymize your assets (if applicable). You can either create an anonymized URL or include an anonymized zip file.
    \end{itemize}

\item {\bf Crowdsourcing and research with human subjects}
    \item[] Question: For crowdsourcing experiments and research with human subjects, does the paper include the full text of instructions given to participants and screenshots, if applicable, as well as details about compensation (if any)? 
    \item[] Answer: \answerNA{}
    \item[] Justification: The work does not involve new crowdsourcing or human-subject experiments.
    \item[] Guidelines:
    \begin{itemize}
        \item The answer \answerNA{} means that the paper does not involve crowdsourcing nor research with human subjects.
        \item Including this information in the supplemental material is fine, but if the main contribution of the paper involves human subjects, then as much detail as possible should be included in the main paper. 
        \item According to the NeurIPS Code of Ethics, workers involved in data collection, curation, or other labor should be paid at least the minimum wage in the country of the data collector. 
    \end{itemize}

\item {\bf Institutional review board (IRB) approvals or equivalent for research with human subjects}
    \item[] Question: Does the paper describe potential risks incurred by study participants, whether such risks were disclosed to the subjects, and whether Institutional Review Board (IRB) approvals (or an equivalent approval/review based on the requirements of your country or institution) were obtained?
    \item[] Answer: \answerNA{}
    \item[] Justification: The work does not involve new human-subject studies requiring IRB or equivalent review.
    \item[] Guidelines:
    \begin{itemize}
        \item The answer \answerNA{} means that the paper does not involve crowdsourcing nor research with human subjects.
        \item Depending on the country in which research is conducted, IRB approval (or equivalent) may be required for any human subjects research. If you obtained IRB approval, you should clearly state this in the paper. 
        \item We recognize that the procedures for this may vary significantly between institutions and locations, and we expect authors to adhere to the NeurIPS Code of Ethics and the guidelines for their institution. 
        \item For initial submissions, do not include any information that would break anonymity (if applicable), such as the institution conducting the review.
    \end{itemize}

\item {\bf Declaration of LLM usage}
    \item[] Question: Does the paper describe the usage of LLMs if it is an important, original, or non-standard component of the core methods in this research? Note that if the LLM is used only for writing, editing, or formatting purposes and does \emph{not} impact the core methodology, scientific rigor, or originality of the research, declaration is not required.
    \item[] Answer: \answerYes{}
    \item[] Justification: The paper describes the VLM backbones used for evaluation and does not use an LLM as a hidden or non-standard component of the frame-selection algorithm.
    \item[] Guidelines:
    \begin{itemize}
        \item The answer \answerNA{} means that the core method development in this research does not involve LLMs as any important, original, or non-standard components.
        \item Please refer to our LLM policy in the NeurIPS handbook for what should or should not be described.
    \end{itemize}

\end{enumerate}